\documentclass[]{jingdong}
\usepackage[toc,page,header]{appendix}

\usepackage{amsmath,amsfonts,bm}

\def\eqref#1{equation~\ref{#1}}
\def\1{\bm{1}}

\DeclareMathAlphabet{\mathsfit}{\encodingdefault}{\sfdefault}{m}{sl}
\SetMathAlphabet{\mathsfit}{bold}{\encodingdefault}{\sfdefault}{bx}{n}

\usepackage{amsthm}
\usepackage{float}

\theoremstyle{plain}

\theoremstyle{definition}

\theoremstyle{remark}

\newtcolorbox{takeaway}{
    colback=black!3, colframe=black!55,
    boxrule=0.6pt, arc=1pt, boxsep=0pt,
    left=6pt, right=6pt, top=4pt, bottom=4pt,
    before skip=6pt, after skip=6pt
}

\AddToHook{cmd/@makecol/before}{%
    \ifnum\value{page}=\getpagerefnumber{fig:intro}\relax
        \raggedbottom
    \fi
}

\title{RL Starts before RL: On Policy Distillation for Better Reinforcement Learning}

\newcommand{\opdrl}{OPD$\rightarrow$RL}
\newcommand{\baserl}{Base$\rightarrow$RL}
\newcommand{\sftrl}{SFT$\rightarrow$RL}

\author[1,2,3,\ddagger]{Shuai Dong}
\author[3,\ddagger]{Yongfu Zhu}
\author[3,\ddagger]{Yuqi Xu}
\author[3,4,\ddagger]{Weichu Xie}
\author[3,4,\ddagger]{Liuwenpu}
\author[3,4,\ddagger]{Ziyue Wang}
\author[3,5,\ddagger]{Kaiwen Tuo}
\author[3,\ddagger]{Congcong Wang}
\author[6,*]{Siyuan Wang}
\author[1,2,*]{Zhongyu Wei}
\author[3,\dagger]{Jiaqi Wang}

\affiliation[1]{Fudan University}
\affiliation[2]{SII}
\affiliation[3]{JD.COM}
\affiliation[4]{Peking University}
\affiliation[5]{The Hong Kong University of Science and Technology}
\affiliation[6]{The Chinese University of Hong Kong}

\contribution[\ddagger]{Equal contribution}
\contribution[*]{Corresponding authors}
\contribution[\dagger]{Project Leader}
\contribution{See the \hyperref[sec:authors]{Authors} section for the complete author list.}

\abstract{
Reinforcement learning (RL) improves reasoning, but its performance depends on the policy from which training begins. We study on-policy distillation (OPD) as a preparation stage for RL and ask whether its benefits extend beyond improvements in the distilled model's initial accuracy. Under shared RL settings, students initialized with OPD reach higher final performance than those trained with direct RL or supervised fine-tuning followed by RL. This advantage can emerge even when OPD produces little immediate improvement in accuracy. Pre-RL Pass@$k$ does not fully explain the benefit: similar or even higher values do not necessarily lead to better performance after RL. Behavioral analyses point to alignment with the teacher's distribution beyond top-1 agreement as a possible explanation. Such alignment may favor higher-quality reasoning paths while retaining alternatives that RL can further refine using outcome feedback. We further examine how trajectory sources and divergence objectives affect the value of distillation for subsequent RL. Standard reverse-KL OPD performs better before RL, but forward-KL OPD overtakes it afterward. Student rollouts outperform teacher rollouts under both objectives before and after RL. These findings highlight the importance of both objective choice and the states receiving supervision for subsequent RL. Our results support OPD as preparation for RL and favor forward KL when OPD is followed by RL in our comparison.
}

\begin{document}
\maketitle

\section{Introduction}
\label{sec:intro}
Reinforcement learning (RL) has become an important approach to improving reasoning in language models~\citep{grpo,deepseekr1,dapo}. By learning from feedback on complete responses, a model can improve its ability to solve tasks through repeated training. Its final performance, however, also depends on the policy from which training begins. The starting policy shapes what the model generates during training and how it responds to feedback. Preparing the model before RL is therefore an important part of the training process. The challenge is that a model's current accuracy does not necessarily tell us how well it will learn afterward.

A common preparation step is supervised fine-tuning (SFT) on solutions generated by a stronger teacher~\citep{kim2016sequence,hsieh2023distilling}. On-policy distillation (OPD) offers another route: the student generates its own responses, and the teacher provides a distribution over possible next tokens along those responses~\citep{agarwal2024gkd,li2026rethinkingopd}. This allows the student to learn from teacher feedback on the prefixes it actually encounters. Although OPD is typically evaluated by the performance of the distilled student, the resulting policy can also serve as a starting point for RL. This raises the central question of our study: does OPD prepare students for better RL, beyond improving their immediate accuracy?

\begin{figure}[t]
    \centering
    \includegraphics[width=\textwidth]{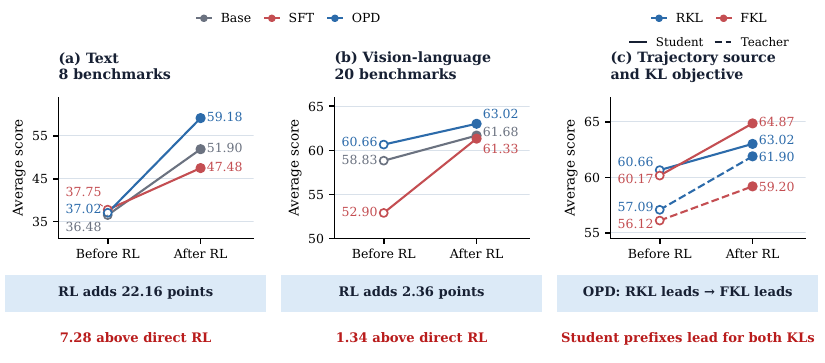}
    \caption{Performance after RL reveals the value of distillation. (a,b) OPD followed by RL improves over the distilled checkpoint and reaches the highest final average. (c) With student-generated prefixes (solid), RKL leads before RL but FKL leads afterward. Student prefixes outperform teacher prefixes, shown as dashed lines, under both objectives. Colors in (c) denote KL objectives; hollow/filled markers throughout denote before/after RL. Panel (c) averages twenty benchmarks (Table~\ref{tab:app-source-objective}).}
    \label{fig:intro}
\end{figure}

We compare direct RL (\baserl{}), SFT followed by RL (\sftrl{}), and OPD followed by RL (\opdrl{}), using the same base student and downstream RL settings within each setting. Students initialized with OPD reach higher final averages than both alternatives (Fig.~\ref{fig:intro}). To understand this advantage, we first examine pre-RL Pass@$k$, which measures whether the model can produce at least one correct response within $k$ attempts and is often used to assess its potential for further RL~\citep{yue2025rlcapacity,wu2025invisible}. A natural explanation is that OPD improves subsequent training by increasing this initial coverage of correct responses. Yet OPD and the base student can begin with nearly overlapping Pass@$k$ curves and reach substantially different performance after RL. Across the ablations, higher initial Pass@$k$ also does not consistently lead to better final results.

These findings motivate a closer look at how OPD changes the student's distribution over reasoning continuations. Agreement with the teacher's most likely token captures only one aspect of this distribution. In our comparisons, SFT can achieve slightly higher top-1 agreement than OPD while showing greater distributional divergence and lower token uncertainty. Strong agreement on the preferred token can therefore coexist with a more concentrated policy, leaving unclear how much flexibility remains for subsequent learning. OPD shows closer distributional alignment while retaining uncertainty over alternatives. Our hypothesis is that this combination helps align the student with the stronger teacher's solution space while preserving potentially high-quality reasoning paths that RL can further reshape. Feedback on complete responses may then help the student adjust the probabilities of these paths and improve task success. This offers a possible explanation for OPD's downstream benefit beyond what initial accuracy or Pass@$k$ alone reveals.

The same distinction between current performance and later learning appears when we change the distillation recipe. We compare student- and teacher-generated trajectories under reverse KL (RKL) and forward KL (FKL), using full teacher distributions in each case. Standard reverse-KL OPD performs better before RL, but forward-KL OPD overtakes it afterward. Student rollouts also outperform teacher rollouts under both objectives before and after RL. These findings highlight both the choice of objective and the states at which teacher supervision is applied when preparing a model for subsequent RL.

These results place OPD within a broader question about preparing models for further learning. Previous work on the SFT-to-RL transition has shown that improving an intermediate model's standalone performance need not improve its usefulness for RL~\citep{kang2025quagmires,malladi2026tailsft}. Our study examines this issue for on-policy teacher transfer, connecting downstream performance with behavioral analysis and choices in the distillation recipe. The evidence supports OPD as a useful preparation stage in the evaluated settings and motivates following the student through RL when assessing the value of distillation.

\noindent Our contributions are:
\begin{itemize}
    \item We evaluate OPD as preparation for RL and show that it yields higher final performance than direct RL and SFT followed by RL under shared downstream training settings.
    \item We offer a possible explanation for OPD's benefit: closer alignment with the teacher's distribution may provide a better solution space for subsequent RL, while retained uncertainty preserves alternative reasoning paths.
    \item We show that forward KL is the better choice when OPD is followed by RL in our comparison, despite reverse KL's stronger standalone distillation performance.
\end{itemize}

Together, these findings support treating OPD as preparation for better RL and choosing distillation recipes by the performance achieved after subsequent training.

\section{Related Work}
\label{sec:related}
\paragraph{Preparing models for RL.}
Work on preparation for RL evaluates intermediate policies by subsequent learning. Behavior injection, exploration-aware fine-tuning, and distillation-based preparation modify supervision for downstream RL~\citep{cen2025behavior,mu2026oxa,li2026sequential}. \citet{kang2025quagmires} show that SFT accuracy can misrank post-RL performance, while TailSFT preserves reward-bearing coverage through data filtering~\citep{malladi2026tailsft}. Studies connect RL improvement to the initial response distribution~\citep{yue2025rlcapacity,wu2025invisible}, prolonged training~\citep{liu2025prorl}, and token entropy~\citep{wang2025entropy}. Our study examines OPD in this context, focusing on downstream gains that initial accuracy and measured Pass@$k$ do not fully explain.

\paragraph{On-policy distillation and objective choice.}
Distillation transfers soft predictions~\citep{hinton2015distilling} or teacher-generated responses~\citep{kim2016sequence,hsieh2023distilling}. MiniLLM uses reverse KL on student rollouts~\citep{gu2024minillm}, while GKD unifies trajectory sources and objectives and also studies joint distillation and RL~\citep{agarwal2024gkd}. Recent work examines supervision stability~\citep{fu2026revisiting}, privileged-information transfer~\citep{yu2026dopd}, distillation objectives~\citep{wu2025rethinkingkl,jin2026entropy}, and student--teacher compatibility~\citep{li2026rethinkingopd}. Simple-OPD uses LoRA warm-up on teacher-generated chain-of-thought before OPD, finding comparable benefits from correct and incorrect teacher traces~\citep{liu2026simpleopd}. Our comparison across twenty benchmarks evaluates distillation objectives and trajectory sources as preparation for RL, showing that the stronger distilled policy need not lead to better post-RL performance.

\section{Experimental Setup}
\label{sec:framework}

\subsection{Models and Preparation before RL}
We compare ways of preparing the same base student for RL. Text experiments use Qwen3-8B~\citep{yang2025qwen3} as the student and Qwen3-32B as the teacher; vision-language (VL) experiments use Qwen3-VL-8B and Qwen3-VL-32B~\citep{bai2025qwen3vl}. Both teachers are RL-trained. In \baserl{}, the base student enters RL directly. In \sftrl{}, supervised next-token prediction on verified teacher-generated responses prepares the student for RL. In \opdrl{}, the student first matches full teacher distributions at its own generated prefixes using reverse KL. Base, SFT, and OPD denote the resulting checkpoints entering RL.

To examine how the preparation recipe affects subsequent learning, we also compare reverse and forward KL on student- and teacher-generated trajectories using the VL model pair. Both trajectory sources receive full teacher distributions, distinguishing teacher-trajectory distillation from SFT with hard token targets. The distillation objectives and the VL OPD RKL/FKL configuration are described in Appendix~\ref{app:experimental-details}; Appendix~\ref{app:source-objective} reports the source--objective comparison.

\subsection{Shared RL Settings and Evaluation}
Teacher transfer and downstream RL use separate reasoning prompt datasets, with sources and sizes detailed in Appendix~\ref{app:experimental-details}. Within each setting, the three pipelines share GRPO-style RL~\citep{grpo}, training data, verifier, optimizer, and sampling configuration. The verifier checks final-answer correctness. Monitoring and reported checkpoints are described in Appendix~\ref{app:experimental-details}.

We evaluate Base, SFT, and OPD before and after RL on eight text and twenty VL benchmarks, with equal task weights within each setting. VL RL mainly uses math-related training data, while evaluation spans STEM and puzzle reasoning, general VQA, instruction following, OCR, and mathematics. Training and evaluation prompts are distinct; complete task lists and checkpoint scores appear in Appendix~\ref{app:full_results}.

\subsection{Behavior before RL}
\label{sec:pre-rl-measurements}
To examine what preparation changes beyond benchmark accuracy, we measure the policies before RL on 1,024 shared diagnostic prompts per setting drawn from the RL training pool. We sample 64 responses per prompt to estimate Pass@$k$~\citep{chen2021codex,lewkowycz2022minerva}, correct-count distributions, and group-outcome probabilities. These diagnostics use a separate sampling protocol from benchmark evaluation and actual RL rollouts.

We examine teacher distributional alignment and retained token uncertainty. Teacher-agreement probes measure Jensen--Shannon (JS) divergence~\citep{lin1991jsd}, top-1 agreement, and Jaccard overlap on separately normalized top-20 prediction records. Each policy supplies its prefixes, so the probes reflect visited contexts and conditional predictions. Token entropy is measured under fixed lexical categories. Diagnostic sample sizes, decoding settings, and metric definitions are detailed in Appendix~\ref{app:metrics}.

\section{OPD as Preparation for RL}
\label{sec:endpoints}

We first test whether OPD prepares students for better performance after RL, including when its immediate effect on accuracy is small. We apply the shared downstream RL procedure to the Base, SFT, and OPD checkpoints and compare their performance before and after training. Table~\ref{tab:pipeline-results} reports benchmark averages, with complete task scores in Appendix~\ref{app:full_results}.

\begin{table}[t]
\centering
\small
\caption{OPD initialization yields the highest final average in both settings. Scores are benchmark averages before and after RL from each starting policy. Stage changes and between-policy margins are differences of the displayed averages, in percentage points.}
\label{tab:pipeline-results}
\setlength{\tabcolsep}{5pt}
\renewcommand{\arraystretch}{1.05}
\begin{tabular*}{\linewidth}{@{\extracolsep{\fill}}llrrr@{}}
\toprule
Setting & Starting policy & Before RL & After RL & Change \\
\midrule
Text & Base & 36.48 & 51.90 & $+15.42$ \\
& SFT & 37.75 & 47.48 & $+9.73$ \\
& OPD & 37.02 & 59.18 & $+22.16$ \\
\midrule
Vision-language & Base & 58.83 & 61.68 & $+2.85$ \\
& SFT & 52.90 & 61.33 & $+8.43$ \\
& OPD & 60.66 & 63.02 & $+2.36$ \\
\bottomrule
\end{tabular*}
\end{table}

\begin{figure}[t]
    \centering
    \includegraphics[width=\textwidth]{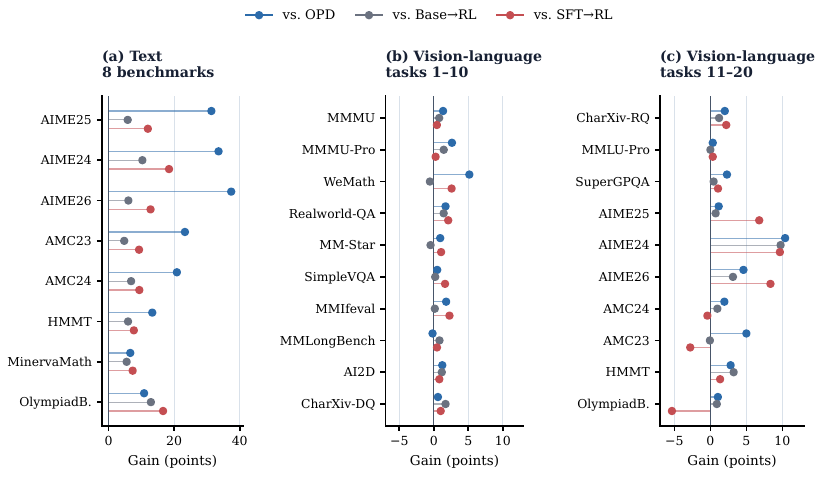}
    \caption{OPD initialization improves final performance across most benchmarks. Each task shows \opdrl{} minus \baserl{} and \sftrl{}, alongside the improvement over the OPD checkpoint before RL. Positive values favor \opdrl{}. All eight text and twenty VL tasks are shown, including ties and regressions; the VL panels follow Appendix~\ref{app:full_results}'s order. Axis scales differ between text and VL.}
    \label{fig:benchmark-gains}
\end{figure}

Initializing RL from OPD gives the highest final average in both settings: 59.18 in text and 63.02 in VL. These scores exceed direct RL by 7.28 and 1.34 points, respectively, and SFT-initialized RL by 11.70 and 1.69 points. The advantage extends across tasks (Fig.~\ref{fig:benchmark-gains}). OPD-initialized RL outperforms both alternatives on all eight text benchmarks; in VL, it exceeds direct RL on 16 of twenty tasks, with one tie, and SFT-initialized RL on 17 of twenty.

The text setting makes the role of preparation particularly clear. OPD changes the initial average only slightly, from 36.48 to 37.02, yet the gap over Base grows from 0.54 points before RL to 7.28 points afterward. SFT starts higher than OPD at 37.75 but finishes lower at 47.48. On AIME24, OPD even starts below Base (24.17 versus 24.90), then reaches 57.71 after RL compared with 47.40 from Base. OPD gains 22.16 points during RL overall. These comparisons show that a small immediate accuracy gain can accompany a much larger downstream advantage, and that the stronger checkpoint before RL need not yield the stronger model afterward.

In VL, OPD leads Base by 1.83 points before RL and retains a 1.34-point advantage afterward. The smaller margin than in text may partly reflect math-focused RL training alongside broader evaluation on twenty benchmarks spanning mathematics, STEM/puzzle reasoning, general VQA, instruction following, and OCR. The 1.34-point average gain is nevertheless notable, with improvements over direct RL on 16 of twenty benchmarks. RL improves the OPD checkpoint by another 2.36 points, with gains on 19 of twenty tasks. RL initialized with OPD achieves the highest final score of 63.02, compared with 61.68 for direct RL and 61.33 for RL initialized with SFT.

\begin{takeaway}
\textbf{Takeaway.} Standard RKL OPD prepares students for higher post-RL performance than direct RL or SFT followed by RL in our comparisons. This benefit can emerge even when the immediate accuracy gain from OPD is small.
\end{takeaway}

\section{Understanding the Benefits of OPD for RL}
\label{sec:support}

The results in Sec.~\ref{sec:endpoints} show that OPD can improve performance after RL even when its immediate accuracy gain is small. We now examine what changes in the policy before RL. We first test whether initial correct-answer coverage explains the downstream advantage, then examine teacher distributional alignment and retained token uncertainty. All diagnostics describe the checkpoints entering RL, using the shared prompt sets in Sec.~\ref{sec:pre-rl-measurements}.

\subsection{Initial Coverage and Subsequent Performance}
\label{sec:initial-coverage}

\begin{figure}[t]
    \centering
    \includegraphics[width=\textwidth]{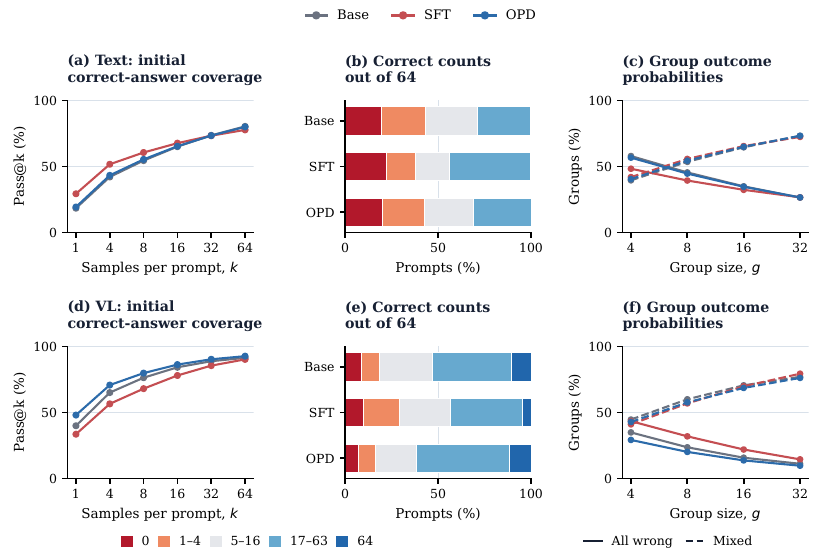}
    \caption{Similar initial coverage can precede different performance after RL. Rows show text and VL. (a,d) Pre-RL Pass@$k$; (b,e) correct-response counts among 64 attempts per prompt; (c,f) estimated all-wrong and mixed-correctness group probabilities. Text Base and OPD have nearly overlapping sampling curves despite their different post-RL scores. Measurements use 1,024 prompts at temperature 0.6; group probabilities are diagnostic estimates computed from the sampled responses.}
    \label{fig:passk}
\end{figure}

A natural explanation is that OPD makes correct responses more accessible, giving RL more successful behavior to reinforce. Pass@$k$ measures whether at least one of $k$ attempts is correct and has been used to assess the potential of a starting policy for RL~\citep{kang2025quagmires,malladi2026tailsft}. We therefore compare the pre-RL Pass@$k$ curves of Base, SFT, and OPD in Fig.~\ref{fig:passk}.

The text comparison shows a limit of this explanation. Base and OPD have nearly overlapping curves: their Pass@1 values are 18.4\% and 19.3\%, and their Pass@64 values are 80.4\% and 80.0\%. Nevertheless, they reach final benchmark averages of 51.90 and 59.18 after RL. Their correct-count distributions and group-outcome probabilities are also similar. For diagnostic groups of eight, the estimated mixed-correctness rates are 53.54\% for Base and 54.41\% for OPD. These summaries reveal little of the eventual performance gap.

Coverage does improve in VL, where OPD raises Pass@1 from 40.0\% to 48.1\% and Pass@64 from 91.4\% to 92.7\% relative to Base. The KL comparison provides another case where coverage alone is insufficient: RKL and FKL have the same pre-RL Pass@64 of 92.68\%, yet FKL finishes 1.85 points higher after RL (Sec.~\ref{sec:interventions}). Thus, increased coverage can accompany stronger downstream performance, but the measured Pass@$k$ values do not fully explain the benefit of preparation across these comparisons.

Similar aggregate coverage does not imply that policies solve the same prompts or follow the same reasoning paths. It also does not specify how probabilities will change during RL. The group diagnostics add information about reward variation: both all-wrong and all-correct groups lack a within-group outcome contrast. They still characterize sampled outcomes before training rather than the policy's subsequent response to optimization. This motivates examining how OPD changes the distribution over continuations.

\subsection{Teacher Alignment beyond Top-1 Agreement}
\label{sec:teacher-alignment}

We compare student and teacher predictions at prefixes visited by each policy, measuring both top-1 agreement and JS divergence between their normalized top-20 records. These metrics capture different aspects of teacher alignment: the preferred token can agree even when the probabilities assigned to alternative tokens differ substantially.

\begin{figure}[t]
    \centering
    \includegraphics[width=\textwidth]{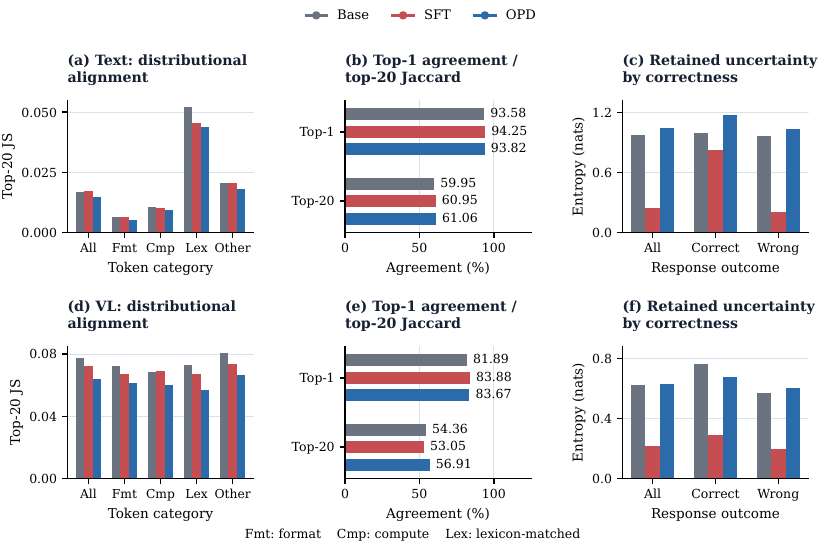}
    \caption{OPD combines closer distributional alignment with retained token uncertainty. Rows show text and VL. (a,d) Top-20 JS overall and by lexical token category; (b,e) top-1 agreement and top-20 Jaccard overlap; (c,f) entropy at lexicon-matched positions, split by response correctness. SFT has slightly higher top-1 agreement than OPD, while OPD has lower JS and higher entropy. Probes use each policy's own prefixes. ``Lex'' denotes the fixed lexicon in Appendix~\ref{app:role-definitions}.}
    \label{fig:teacher-behavior}
\end{figure}

OPD reduces the recorded distributional divergence from the teacher in both settings (Fig.~\ref{fig:teacher-behavior}). Mean top-20 JS falls from 0.01701 for Base to 0.01468 for OPD in text, and from 0.07734 to 0.06431 in VL. This change is visible even in text, where the initial accuracy and Pass@$k$ curves are close. OPD therefore changes teacher alignment without requiring a large increase in initial correct-answer frequency.

The comparison with SFT shows why top-1 agreement alone is incomplete. Text SFT has higher top-1 agreement than OPD, 94.25\% versus 93.82\%, but also higher JS, 0.01720 versus 0.01468. The same pattern appears in VL: SFT has 83.88\% top-1 agreement compared with OPD's 83.67\%, while its JS is higher, 0.07208 versus 0.06431. Agreement on the most likely token therefore does not establish closeness of the surrounding distribution.

\subsection{Retaining Uncertainty for Subsequent Learning}
\label{sec:retained-uncertainty}

Closer teacher alignment leaves open how much uncertainty remains over alternative continuations. We examine token entropy at positions matching a fixed reasoning-word lexicon, with separate summaries for correct and incorrect responses (Fig.~\ref{fig:teacher-behavior}, right column). OPD retains entropy comparable to Base: 1.048 versus 0.971 in text and 0.629 versus 0.622 in VL. SFT has lower values, 0.242 and 0.214, respectively. The OPD--SFT difference also appears within both correctness groups, so it is not explained solely by the proportion of correct responses.

Taken together, the alignment and entropy measurements suggest a possible explanation for OPD's benefit. Teacher feedback may shift probability toward higher-quality continuations while retaining alternatives that remain useful for subsequent learning. Outcome-based RL can then adjust the probabilities of complete responses using task feedback. In this account, OPD helps shape the choices available to the student, while RL refines how often those choices lead to successful answers. This connects the downstream advantage to changes in the policy distribution that initial accuracy or Pass@$k$ alone does not describe.

\begin{takeaway}
\textbf{Takeaway.} Our analysis of standard RKL OPD suggests that closer alignment with the teacher's distribution may provide a better solution space for subsequent RL, with retained uncertainty preserving alternatives for further refinement.
\end{takeaway}

\section{Choosing the Distillation Recipe for Subsequent RL}
\label{sec:interventions}

% Queue the KL figure before the table so it appears on page 8.
\begin{figure}[!t]
    \centering
    \includegraphics[width=\textwidth]{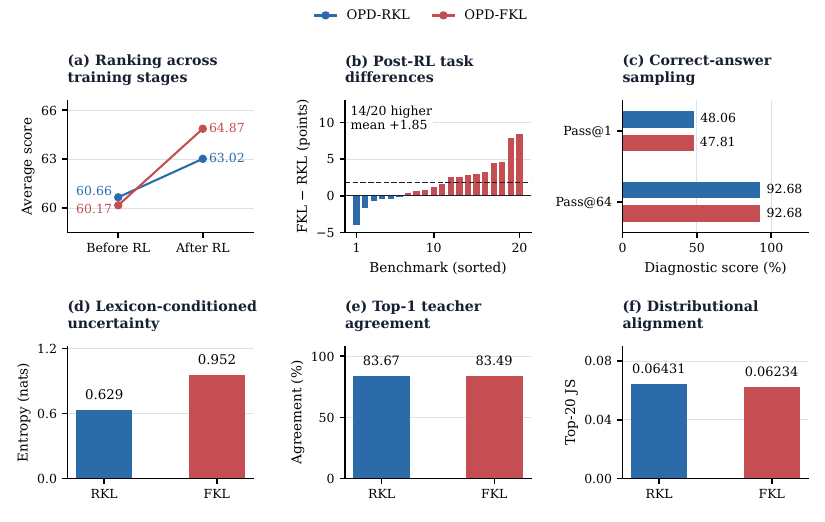}
    \caption{FKL prepares a stronger starting policy for RL despite similar initial coverage. Both variants use student-generated trajectories. (a) Benchmark averages before and after RL; (b) all twenty post-RL task differences, sorted, with a dashed mean; (c--f) pre-RL Pass@$k$, lexicon-conditioned entropy, top-1 teacher agreement, and top-20 JS. FKL has higher entropy and lower JS while reaching higher performance after RL.}
    \label{fig:kl-sequence}
\end{figure}

We next ask how the distillation recipe affects preparation for RL. We compare RKL and FKL across the same twenty benchmarks using student- and teacher-generated trajectories, Qwen3-VL models, and shared RL settings. Both trajectory sources receive full teacher distributions; teacher-trajectory distillation (TFD) therefore differs from SFT with hard token targets. Table~\ref{tab:interventions} reports performance before and after RL, with configurations in Appendix~\ref{app:experimental-details}.

\begin{table}[t]
\centering
\small
\caption{Student rollouts outperform teacher rollouts under both KL objectives before and after RL. Scores average twenty benchmarks for Qwen3-VL-8B. All four variants use full teacher distributions. Changes are computed before rounding; each row tracks a distillation configuration through its subsequent RL stage.}
\label{tab:interventions}
\setlength{\tabcolsep}{4pt}
\begin{tabular*}{\linewidth}{@{\extracolsep{\fill}}llrrr@{}}
\toprule
Distillation prefixes & Objective & Before RL & After RL & Change \\
\midrule
Student (OPD) & RKL & 60.66 & 63.02 & $+2.36$ \\
Student (OPD) & FKL & 60.17 & 64.87 & $+4.70$ \\
\midrule
Teacher (TFD) & RKL & 57.09 & 61.90 & $+4.82$ \\
Teacher (TFD) & FKL & 56.12 & 59.20 & $+3.08$ \\
\bottomrule
\end{tabular*}
\end{table}

With student-generated trajectories, the objective ranking reverses after RL. Standard RKL OPD starts at 60.66, above FKL's 60.17, but finishes at 63.02, below FKL's 64.87. FKL gains 4.70 points during RL compared with RKL's 2.36, and its final advantage spans 14 of twenty benchmarks (Fig.~\ref{fig:kl-sequence}). Selecting by immediate distillation performance would favor RKL, whereas selecting by performance after RL favors FKL.

Distillation on teacher rollouts yields lower scores than distillation on student rollouts under both FKL and RKL, before and after RL. In the twenty-benchmark average after RL, student rollouts lead by 5.67 points under FKL and 1.12 points under RKL. Both sources use full teacher distributions, so this comparison highlights the importance of the states at which supervision is applied: teacher feedback on prefixes visited by the student provides better preparation for RL in our comparison. Fig.~\ref{fig:intro}(c) contrasts the two trajectory sources, with the full comparison in Appendix~\ref{app:source-objective}.

The OPD comparison also shows why initial Pass@$k$ is insufficient to choose between the objectives. FKL reaches a Pass@1 of 47.81\%, slightly below RKL's 48.06\%, while both achieve a Pass@64 of 92.68\%. Despite this similar initial coverage, FKL reaches higher performance after RL. FKL also retains higher entropy at positions matching the reasoning lexicon, at 0.952 compared with 0.629 for RKL, and achieves lower JS, at 0.06234 compared with 0.06431. Its agreement with the teacher's most likely token is slightly lower, at 83.49\% compared with 83.67\% for RKL. This pattern is consistent with the explanation in Sec.~\ref{sec:support}: closer distributional alignment can coexist with retained uncertainty over alternatives for subsequent learning.

FKL penalizes insufficient probability on continuations supported by the teacher, whereas RKL penalizes probability assigned to continuations the teacher disfavors~\citep{agarwal2024gkd,gu2024minillm}. Together with the advantage of student rollouts under both objectives, our results suggest that preparation for RL depends on both how teacher feedback shapes the policy and which states receive that feedback.

\begin{takeaway}
\textbf{Takeaway.} FKL is better for OPD followed by RL despite RKL's higher pre-RL score. Student rollouts outperform teacher rollouts under both objectives in our comparison.
\end{takeaway}

\section{Discussion}
\label{sec:discussion}

Our results support treating OPD as part of the design of RL. A preparation stage can yield little immediate accuracy improvement yet lead to substantially better performance after training. The objective comparison makes the same point at the level of the distillation recipe: the strongest distilled checkpoint need not be the best starting policy for RL. The advantage of student rollouts under both objectives supports evaluating distillation choices by subsequent RL performance.

Our behavioral analysis of standard RKL OPD offers a possible explanation for its value as preparation for RL. In our distributional probes, OPD shows closer teacher alignment beyond top-1 agreement while retaining uncertainty over alternatives, even when initial accuracy and measured Pass@$k$ change little. We hypothesize that this combination provides a better solution space for subsequent RL, understood as the reasoning continuations available to the policy and the probabilities assigned to them. Teacher alignment may favor higher-quality continuations, while retained uncertainty may preserve alternatives for exploration and refinement. RL can then use outcome feedback on complete responses to adjust these probabilities. In this account, OPD's benefit lies partly in shaping how the policy can learn afterward, beyond improving what it already answers correctly.

Our analysis characterizes teacher distributional alignment and retained uncertainty jointly. Future work could further disentangle their individual contributions and examine how alternative continuations are reinforced during RL. Such studies could guide the design of distillation objectives that improve subsequent learning, as well as criteria for selecting starting policies before running RL.

\section{Conclusion}
\label{sec:conclusion}
Our results show that OPD improves final performance over direct RL and SFT followed by RL, even when its immediate accuracy gain is small. Its benefit extends beyond what pre-RL Pass@$k$ reveals, with closer alignment to the teacher's distribution and retained uncertainty offering a possible explanation. This focus on subsequent learning also informs the distillation recipe: FKL is preferable for OPD followed by RL, and student rollouts outperform teacher rollouts under both objectives in our comparison. Together, these findings support OPD as preparation for better RL and highlight the importance of shaping starting policies for the learning that follows.

\subsection*{AI use statement}
We used an AI assistant to polish the writing and improve the clarity and readability of the manuscript. The authors are responsible for reviewing the AI-assisted material and for the final claims, analyses, and conclusions.

\clearpage
\section{Authors}
\label{sec:authors}

\textbf{Full Author List}\\[0.5em]
Shuai Dong$^{1,2,3,\ddagger}$,
Yongfu Zhu$^{3,\ddagger}$,
Yuqi Xu$^{3,\ddagger}$,
Weichu Xie$^{3,4,\ddagger}$,
Liuwenpu$^{3,4,\ddagger}$,
Ziyue Wang$^{3,4,\ddagger}$,
Kaiwen Tuo$^{3,5,\ddagger}$,
Congcong Wang$^{3,\ddagger}$,
Siyuan Wang$^{6,*}$,
Wenqi Shao$^{3}$,
Shuai Yang$^{7}$,
Ji Zhao$^{3}$,
Caoyuan Ma$^{3,8}$,
Wenzheng Chang$^{3,7}$,
Taiqiang Wu$^{3,9}$,
Xinlei Yu$^{3,6}$,
Hongrui Wu$^{3,10}$,
Xiaoxuan He$^{11}$,
Fangke Chen$^{2,11}$,
Dianyi Wang$^{2,3}$,
Kanghui Tian$^{13}$,
Sirry Chen$^{12}$,
Xingyu Liu$^{12}$,
Xiangnan Wu$^{13}$,
Jiawei Guo$^{13}$,
Haowen Hou$^{2,3,7}$,
LingHan Chen$^{1}$,
Zhongyu Wei$^{1,2,*}$,
Jiaqi Wang$^{3,\dagger}$

\vspace{1em}
\textbf{Affiliations}\\[0.5em]
\begin{tabular}{@{}rl@{}}
$^{1}$ & Fudan University \\
$^{2}$ & SII \\
$^{3}$ & JD.COM \\
$^{4}$ & Peking University \\
$^{5}$ & The Hong Kong University of Science and Technology \\
$^{6}$ & The Chinese University of Hong Kong \\
$^{7}$ & Shanghai Jiao Tong University \\
$^{8}$ & The University of Tokyo \\
$^{9}$ & The University of Hong Kong \\
$^{10}$ & Stanford University \\
$^{11}$ & Zhejiang University \\
$^{12}$ & China University of Geosciences \\
$^{13}$ & Institute of Automation, Chinese Academy of Sciences
\end{tabular}

\vspace{1em}
$^{\ddagger}$Equal contribution. \quad
$^{*}$Corresponding authors. \quad
$^{\dagger}$Project Leader.

\clearpage
\bibliographystyle{plainnat}
\bibliography{cite}

\appendix
% Supplementary material follows the preparation-for-RL narrative.
\clearpage
\begingroup
\raggedbottom
\setlength{\floatsep}{10pt plus 2pt minus 2pt}
\setlength{\textfloatsep}{12pt plus 2pt minus 2pt}
\setlength{\intextsep}{10pt plus 2pt minus 2pt}
% Preparation and subsequent RL are documented separately; saved settings are retained.
\section{Preparation, RL, and Evaluation Protocols}
\label{app:experimental-details}

The experiments evaluate preparation by the performance reached after RL. This appendix defines the starting policies and data sources and documents the VL OPD objective-comparison configuration. Appendix~\ref{app:full_results} reports the complete benchmark results, Appendix~\ref{app:metrics} defines the diagnostics before RL, and Appendix~\ref{app:ablations} compares trajectory sources and distillation objectives. Diagnostic scoring appears in Appendix~\ref{app:response-processing}.

\subsection{Models, Data, and Evaluation Suites}

The text setting pairs a Qwen3-8B student with a Qwen3-32B teacher. The vision-language setting pairs Qwen3-VL-8B with Qwen3-VL-32B. Both teachers have undergone RL. Within each setting, Base, SFT, and OPD originate from the same student, and their subsequent RL runs share the training data, verifier, optimizer, and sampling configuration.

Text SFT and OPD use 100K OpenMathInstruct prompts~\citep{toshniwal2024openmath}. SFT learns from verified teacher responses, while OPD queries the teacher along responses generated by the student. Text RL uses 17K DAPO-Math prompts. VL OPD uses 100K OpenMMReasoner prompts~\citep{zhang2025openmmreasoner}, and VL SFT uses teacher-generated solutions. VL RL uses 16K prompts drawn from DeepMath and MMK12/VisionR1~\citep{he2025deepmath,huang2025visionr1}. OlympiadBench~\citep{he2024olympiadbench} is used only for evaluation.

\begin{table}[H]
\centering
\small
\caption{Data and evaluation for preparation and subsequent RL. The benchmark suites are distinct from the training prompts. The VL suite also evaluates the trajectory-source and objective comparisons.}
\label{tab:app-data-overview}
\setlength{\tabcolsep}{4pt}
\renewcommand{\arraystretch}{1.10}
\begin{tabular*}{\linewidth}{@{\extracolsep{\fill}}p{0.19\linewidth}p{0.32\linewidth}p{0.37\linewidth}@{}}
\toprule
Component & Text & Vision-language \\
\midrule
Student / teacher & Qwen3-8B / Qwen3-32B & Qwen3-VL-8B / Qwen3-VL-32B \\
OPD prompts & OpenMathInstruct, 100K & OpenMMReasoner, 100K \\
RL prompts & DAPO-Math, 17K & DeepMath, MMK12/VisionR1, 16K in total \\
Evaluation & Eight benchmarks & Twenty benchmarks \\
Diagnostic prompts & 1,024 from the RL pool & 1,024 from the RL pool \\
\bottomrule
\end{tabular*}
\end{table}

The benchmark suites measure task performance, while the diagnostic prompts support analysis of the policies entering RL. Each benchmark receives equal weight in its setting's average. Complete task lists and scores are given in Appendix~\ref{app:full_results}; diagnostic sample sizes and decoding settings are specified in Appendix~\ref{app:diagnostic-sampling}.

\subsection{Starting Policies and Distillation Objectives}

Base enters RL directly. SFT first learns to predict the tokens of verified teacher responses. With teacher policy $\pi_T$, student policy $\pi_\theta$, prompt $x$, and teacher response $y^T$, its objective is
\begin{equation}
\mathcal L_{\mathrm{SFT}}
=-\mathbb E_{(x,y^T)}\frac{1}{|y^T|}
\sum_t\log\pi_\theta(y_t^T\mid x,y^T_{<t}).
\end{equation}
OPD instead uses the teacher's full next-token distribution at prefixes generated by the current student. For prefix $s$, write $p_s(a)=\pi_\theta(a\mid s)$ and $q_s(a)=\pi_T(a\mid s)$. We consider
\begin{equation}
\mathcal L_{\mathrm{RKL}}(s)=D_{\mathrm{KL}}(p_s\Vert q_s),
\qquad
\mathcal L_{\mathrm{FKL}}(s)=D_{\mathrm{KL}}(q_s\Vert p_s).
\end{equation}
The main OPD checkpoints use reverse KL. The objective comparison also applies forward KL at student-generated prefixes. Teacher-trajectory distillation, denoted TFD, evaluates the same two losses at teacher-generated prefixes and retains full teacher distributions as targets. Thus, trajectory source specifies where supervision is applied, and the KL objective specifies how the student matches the teacher there. SFT uses hard token targets on teacher responses.

Both KL objectives are minimized at $p_s=q_s$ for an unrestricted local distribution. Their empirical differences arise with the finite student, the visited prefixes, and the optimization procedure. Training uses full teacher distributions; the top-20 records in Appendix~\ref{app:local-definitions} are used only for diagnostic probes.

\subsection{Training Configuration and Reported Checkpoints}

Table~\ref{tab:app-kl-configuration} specifies the VL OPD-RKL and OPD-FKL pipelines and reports the stage-specific seeds. The OPD stage uses bfloat16. This configuration applies to the VL OPD objective comparison.

\begin{table}[H]
\centering
\small
\caption{Configuration of OPD-RKL, OPD-FKL, and their subsequent RL runs. The student is Qwen3-VL-8B and the teacher is Qwen3-VL-32B. Both variants use the same downstream RL configuration.}
\label{tab:app-kl-configuration}
\setlength{\tabcolsep}{3pt}
\renewcommand{\arraystretch}{1.06}
\begin{minipage}[t]{0.48\linewidth}
\textbf{OPD stage}\par\vspace{2pt}
\begin{tabular*}{\linewidth}{@{\extracolsep{\fill}}p{0.57\linewidth}p{0.37\linewidth}@{}}
\toprule
Setting & Value \\
\midrule
Objective & RKL or FKL \\
Training data size & 100K prompts \\
Epochs & 1 \\
Global batch size & 1,024 \\
Micro-batch size & 1 \\
Learning rate & $10^{-5}$ \\
Learning-rate schedule & Cosine; 5\% warmup \\
Minimum learning rate & $10^{-6}$ \\
Maximum prompt / response tokens & 4,096 / 16,384 \\
Precision & bfloat16 \\
Seed & 42 \\
\bottomrule
\end{tabular*}
\end{minipage}
\hfill
\begin{minipage}[t]{0.48\linewidth}
\textbf{RL stage}\par\vspace{2pt}
\begin{tabular*}{\linewidth}{@{\extracolsep{\fill}}p{0.57\linewidth}p{0.37\linewidth}@{}}
\toprule
Setting & Value \\
\midrule
Algorithm & GRPO \\
Training data size & 16K prompts \\
PPO epochs per update & 1 \\
Actor global batch size & 64 \\
Rollout batch size & 256 prompts \\
Responses per prompt & 10 \\
Learning rate & $10^{-6}$ \\
Maximum prompt / response tokens & 4,096 / 16,384 \\
Sampling temperature / top-$p$ & 1.0 / 1.0 \\
Actor KL-loss coefficient & 0.01 \\
Seed & 1 \\
Validation / checkpoint interval & 5 / 10 steps \\
\bottomrule
\end{tabular*}
\end{minipage}
\end{table}

Downstream RL uses GRPO-style updates with a verifier that checks the final boxed answer. Training stops when monitoring indicates training instability, characterized by a sharp rise or fall in policy entropy over a few steps, a sharp drop in reward, and the onset of incoherent outputs in the training logs. The same monitoring criteria apply within each setting.

The benchmark tables report checkpoints from these runs. Our comparisons focus on post-RL benchmark performance under shared downstream settings; preparation cost and total training compute are outside the comparison metric.

\subsection{Score Aggregation and Uncertainty}

Benchmark scores are reported as percentages. For each starting policy, the score before RL measures the immediate result of preparation, the score after RL measures downstream performance, and their difference measures improvement during RL. Main-pipeline margins use the displayed two-decimal averages, as stated in Table~\ref{tab:pipeline-results}. The objective comparison uses unrounded averages before computing changes. Task-level differences use the individual benchmark scores.

Appendix~\ref{app:full_results} reports the benchmark scores, and Appendix~\ref{app:metrics} specifies diagnostic decoding and aggregation. Bootstrap intervals quantify variation across the sampled diagnostic prompts for fixed checkpoints.

\clearpage
% All task scores and their order are preserved from the saved benchmark tables.
\section{Complete Performance before and after RL}
\label{app:full_results}

The tables report every benchmark score before and after RL from Base, SFT, and OPD. Scores are percentages, and each task receives equal weight in its setting's average. The paired columns distinguish immediate preparation gains from downstream performance and supply the differences in Fig.~\ref{fig:benchmark-gains}.

OPD-initialized RL achieves the highest final score on all eight tasks. Its initial average exceeds Base by only 0.54 points, while the final margin reaches 7.28 points. This includes AIME24, where OPD initially trails Base.

\begin{table}[H]
\centering
\small
\setlength{\tabcolsep}{3pt}
\renewcommand{\arraystretch}{1.05}
\caption{All eight text benchmarks before and after RL from Base, SFT, and OPD. The OPD column uses reverse-KL distillation.}
\label{tab:appendix-text-results}
\begin{tabular*}{\linewidth}{@{\extracolsep{\fill}}lrrrrrr@{}}
\toprule
& \multicolumn{2}{c}{\baserl{}} & \multicolumn{2}{c}{\sftrl{}} & \multicolumn{2}{c}{\opdrl{}} \\
\cmidrule(lr){2-3}\cmidrule(lr){4-5}\cmidrule(lr){6-7}
Benchmark & Before & After & Before & After & Before & After \\
\midrule
AIME25         & 19.90 & 45.83 & 25.00 & 39.69 & 20.31 & 51.67 \\
AIME24         & 24.90 & 47.40 & 25.62 & 39.27 & 24.17 & 57.71 \\
AIME26         & 14.27 & 47.92 & 21.98 & 41.15 & 16.56 & 53.96 \\
AMC23          & 66.56 & 86.25 & 64.84 & 81.72 & 67.73 & 91.02 \\
AMC24          & 44.44 & 58.40 & 44.44 & 55.90 & 44.44 & 65.28 \\
HMMT           & 11.15 & 17.60 & 11.67 & 15.83 & 10.21 & 23.54 \\
MinervaMath    & 55.88 & 58.82 & 56.99 & 56.99 & 57.72 & 64.34 \\
OlympiadBench  & 54.75 & 52.97 & 51.48 & 49.26 & 55.04 & 65.88 \\
\midrule
Average & 36.48 & 51.90 & 37.75 & 47.48 & 37.02 & 59.18 \\
\bottomrule
\end{tabular*}
\end{table}

OPD improves on 19 of twenty tasks during RL. Its final score exceeds direct RL on 16 tasks with one tie and exceeds SFT-initialized RL on 17 tasks. All remaining ties and regressions are retained below.

\begin{table}[H]
\centering
\small
\setlength{\tabcolsep}{3pt}
\renewcommand{\arraystretch}{1.05}
\caption{All twenty VL benchmarks before and after RL. The task order matches the VL panels of Fig.~\ref{fig:benchmark-gains}, and the average includes every row.}
\label{tab:appendix-vl-results}
\begin{tabular*}{\linewidth}{@{\extracolsep{\fill}}lrrrrrr@{}}
\toprule
& \multicolumn{2}{c}{\baserl{}} & \multicolumn{2}{c}{\sftrl{}} & \multicolumn{2}{c}{\opdrl{}} \\
\cmidrule(lr){2-3}\cmidrule(lr){4-5}\cmidrule(lr){6-7}
Benchmark & Before & After & Before & After & Before & After \\
\midrule
MMMU              & 65.78 & 66.89 & 64.00 & 67.22 & 66.33 & 67.67 \\
MMMU-Pro          & 52.20 & 53.27 & 49.97 & 54.45 & 52.08 & 54.71 \\
WeMath            & 52.95 & 62.67 & 48.38 & 59.52 & 56.95 & 62.10 \\
Realworld-QA      & 70.98 & 72.03 & 70.46 & 71.37 & 71.76 & 73.46 \\
MM-Star           & 71.20 & 72.60 & 68.73 & 71.07 & 71.20 & 72.13 \\
SimpleVQA         & 53.48 & 55.01 & 53.19 & 53.58 & 54.72 & 55.21 \\
MMIfeval          & 60.99 & 63.40 & 60.82 & 61.28 & 61.76 & 63.55 \\
MMLongBench-Doc   & 48.30 & 48.95 & 47.20 & 49.31 & 49.95 & 49.77 \\
AI2D              & 85.07 & 84.46 & 83.68 & 84.82 & 84.39 & 85.62 \\
CharXiv-DQ        & 87.42 & 87.35 & 87.15 & 88.05 & 88.45 & 89.05 \\
CharXiv-RQ        & 46.30 & 48.00 & 45.40 & 47.00 & 47.20 & 49.20 \\
MMLU-Pro          & 68.08 & 68.41 & 64.01 & 68.08 & 68.08 & 68.41 \\
SuperGPQA         & 51.43 & 52.53 & 47.21 & 51.92 & 50.68 & 52.98 \\
AIME25            & 38.44 & 49.17 & 31.77 & 43.12 & 48.75 & 49.90 \\
AIME24            & 49.27 & 54.69 & 31.35 & 54.79 & 54.06 & 64.44 \\
AIME26            & 48.54 & 55.00 & 28.65 & 49.79 & 53.54 & 58.13 \\
AMC24             & 57.78 & 61.32 & 47.99 & 62.71 & 60.35 & 62.29 \\
AMC23             & 82.42 & 87.34 & 66.41 & 90.07 & 82.27 & 87.27 \\
HMMT              & 25.10 & 27.81 & 11.35 & 29.69 & 28.23 & 31.04 \\
OlympiadBench     & 60.83 & 62.61 & 50.30 & 68.84 & 62.46 & 63.50 \\
\midrule
Average  & 58.83 & 61.68 & 52.90 & 61.33 & 60.66 & 63.02 \\
\bottomrule
\end{tabular*}
\end{table}

\clearpage
% Saved diagnostics describe the policies entering RL.
\section{Initial Coverage, Teacher Alignment, and Retained Uncertainty}
\label{app:metrics}

This appendix defines the measurements used to examine what preparation changes before RL. We first describe the diagnostic samples, then report correct-response coverage, agreement with the teacher's distribution, and token uncertainty. These measurements support the analysis in Sec.~\ref{sec:support} and the OPD objective comparison in Sec.~\ref{sec:interventions}.

\subsection{Diagnostic Prompts and Sampling}
\label{app:diagnostic-sampling}

For each setting, all compared policies are evaluated on the same 1,024-prompt subset of the RL training pool. We measure the checkpoints entering RL, separately from the benchmark evaluation in Appendix~\ref{app:full_results}.

Generation uses temperature 0.6 and top-$p=0.95$, with limits of 4,096 prompt tokens and 16,384 response tokens. An instruction to provide a boxed answer is appended when the prompt does not already contain one. Coverage uses 64 responses per prompt. Entropy and teacher probes use a separate generation run with one response per prompt. Table~\ref{tab:app-diagnostic-sampling} specifies the sample and aggregation rule for each measurement.

\begin{table}[H]
\centering
\small
\caption{Samples used for diagnostics before RL. The shared prompt identities allow comparisons across policies, while each diagnostic uses its specified response pool and weighting.}
\label{tab:app-diagnostic-sampling}
\setlength{\tabcolsep}{4pt}
\renewcommand{\arraystretch}{1.10}
\begin{tabular*}{\linewidth}{@{\extracolsep{\fill}}p{0.28\linewidth}p{0.65\linewidth}@{}}
\toprule
Measurement & Samples and aggregation \\
\midrule
Pass@$k$ and group outcomes & 64 responses for each of 1,024 prompts; compute per-prompt values, then average over prompts. \\
Teacher distribution probes & First 512 prompt identities from each policy's separate one-response run; first 2,048 generated positions with stride four; average over matched positions. \\
Token entropy & One separate response for each prompt; average over token positions within lexical categories and response correctness groups. \\
\bottomrule
\end{tabular*}
\end{table}

\subsection{Correct-Response Coverage and Group Outcomes}
\label{app:passk-definition}

For prompt $x$, let $n=64$ and let $c_x$ denote the number of responses marked correct by the diagnostic verifier. Pass@$k$ estimates the probability that at least one of $k$ attempts succeeds. For $1\leq k,g\leq n$, we use the finite-sample estimators
\begin{equation}
\label{eq:app-passk}
\widehat{\mathrm{Pass@}k}(x)
=1-\frac{\binom{n-c_x}{k}}{\binom{n}{k}},
\qquad
\widehat p_{0,g}(x)
=\frac{\binom{n-c_x}{g}}{\binom{n}{g}},
\end{equation}

where $\binom{a}{b}=0$ for $b>a$. The quantity $\widehat p_{0,g}$ is the probability that a group of $g$ distinct responses drawn uniformly without replacement from the 64 sampled responses for that prompt contains no correct answer. Each statistic is computed within a prompt and then averaged over all 1,024 prompts. The correct-count histograms in Fig.~\ref{fig:passk} count prompts in bins $c_x=0$, $1$--$4$, $5$--$16$, $17$--$63$, and $64$.

The corresponding probabilities of an all-correct group and a group containing both correct and incorrect responses are
\begin{equation}
\label{eq:app-mixed-groups}
\widehat p_{1,g}(x)=\frac{\binom{c_x}{g}}{\binom{n}{g}},
\qquad
\widehat p_{\mathrm{mix},g}(x)=1-\widehat p_{0,g}(x)-\widehat p_{1,g}(x).
\end{equation}

These estimates describe response availability under diagnostic sampling. Mixed groups contain a correctness contrast, while all-correct and all-wrong groups do not. The probabilities are computed from sampled responses at temperature 0.6 and summarize diagnostic group outcomes.

\begin{table}[H]
\centering
\small
\caption{Pass@$k$ before RL, reported as percentages and averaged over the shared diagnostic prompts. OPD-RKL and OPD-FKL are the two student-trajectory variants used in Sec.~\ref{sec:interventions}.}
\label{tab:app-passk}
\setlength{\tabcolsep}{3pt}
\renewcommand{\arraystretch}{1.08}
\begin{tabular*}{\linewidth}{@{\extracolsep{\fill}}lrrrrrr@{}}
\toprule
Policy & Pass@1 & Pass@4 & Pass@8 & Pass@16 & Pass@32 & Pass@64 \\
\midrule
Text Base & 18.41 & 42.08 & 54.49 & 65.02 & 73.56 & 80.37 \\
Text SFT & 29.33 & 51.69 & 60.61 & 67.67 & 73.33 & 77.73 \\
Text OPD & 19.33 & 43.29 & 55.38 & 65.36 & 73.42 & 79.98 \\
\midrule
VL Base & 39.96 & 65.09 & 76.35 & 84.25 & 88.85 & 91.41 \\
VL SFT & 33.56 & 56.56 & 68.05 & 78.04 & 85.51 & 90.23 \\
VL OPD-RKL & 48.06 & 70.79 & 79.84 & 86.30 & 90.32 & 92.68 \\
VL OPD-FKL & 47.81 & 70.99 & 80.19 & 86.77 & 90.69 & 92.68 \\
\bottomrule
\end{tabular*}
\end{table}

The text results illustrate why initial coverage does not fully explain later performance. Base and OPD have Pass@64 values of 80.37\% and 79.98\%, yet their final benchmark averages are 51.90 and 59.18. Text SFT has the highest Pass@1 at 29.33\% but the lowest final benchmark average among the three starting policies. In VL, OPD improves coverage over Base, while OPD-RKL and OPD-FKL have identical Pass@64 despite different scores after RL. Coverage therefore provides useful information about the starting policy without determining its downstream ranking.

\begin{table}[H]
\centering
\small
\caption{Estimated probability of an all-wrong group, in percent. Values use Eq.~\ref{eq:app-passk}, with groups drawn without replacement from each prompt's 64-response diagnostic sample.}
\label{tab:app-all-wrong-groups}
\setlength{\tabcolsep}{3pt}
\renewcommand{\arraystretch}{1.08}
\begin{tabular*}{\linewidth}{@{\extracolsep{\fill}}lrrrr@{}}
\toprule
Policy & $g=4$ & $g=8$ & $g=16$ & $g=32$ \\
\midrule
Text Base & 57.92 & 45.51 & 34.98 & 26.44 \\
Text SFT & 48.31 & 39.39 & 32.33 & 26.67 \\
Text OPD & 56.71 & 44.62 & 34.64 & 26.58 \\
\midrule
VL Base & 34.91 & 23.65 & 15.75 & 11.15 \\
VL SFT & 43.44 & 31.95 & 21.96 & 14.49 \\
VL OPD-RKL & 29.21 & 20.16 & 13.70 & 9.68 \\
VL OPD-FKL & 29.01 & 19.81 & 13.23 & 9.31 \\
\bottomrule
\end{tabular*}
\end{table}

\begin{table}[H]
\centering
\small
\caption{Estimated probability of a group containing both correct and incorrect responses, in percent. Values use Eq.~\ref{eq:app-mixed-groups}. The VL OPD row is the standard RKL checkpoint.}
\label{tab:app-mixed-groups}
\setlength{\tabcolsep}{3pt}
\begin{tabular*}{\linewidth}{@{\extracolsep{\fill}}lrrrr@{}}
\toprule
Policy & $g=4$ & $g=8$ & $g=16$ & $g=32$ \\
\midrule
Text Base & 39.53 & 53.54 & 64.55 & 73.27 \\
Text SFT & 41.95 & 55.73 & 65.46 & 72.44 \\
Text OPD & 40.48 & 54.41 & 64.95 & 73.26 \\
VL Base & 44.67 & 60.00 & 70.57 & 77.04 \\
VL SFT & 41.03 & 56.88 & 69.81 & 79.39 \\
VL OPD & 42.82 & 57.67 & 68.59 & 76.19 \\
\bottomrule
\end{tabular*}
\end{table}

Tables~\ref{tab:app-all-wrong-groups} and~\ref{tab:app-mixed-groups} provide the group probabilities shown in Fig.~\ref{fig:passk}. For each group size, the all-correct probability is the remainder after subtracting the other two probabilities from 100\%. With eight responses, text Base and OPD have mixed-group rates of 53.54\% and 54.41\%, providing another similar initial summary despite their different final scores.

To quantify uncertainty in Pass@$k$ differences, we resample 1,024 prompt identities with replacement 5,000 times using NumPy's default random generator with seed 715. All responses belonging to a prompt and the pairing between policies are retained. Each bootstrap statistic averages the per-prompt differences in Eq.~\ref{eq:app-passk}. The 2.5th and 97.5th percentiles give the 95\% interval~\citep{efron1979bootstrap}. These intervals quantify variation across diagnostic prompts for the paired Pass@$k$ differences reported in Table~\ref{tab:passk_bootstrap}.

\begin{table}[H]
\centering
\footnotesize
\caption{Paired Pass@$k$ differences in percentage points, with 95\% intervals from resampling diagnostic prompts. Each resample preserves prompt pairing across policies.}
\label{tab:passk_bootstrap}
\setlength{\tabcolsep}{3pt}
\renewcommand{\arraystretch}{1.08}
\begin{tabular*}{\linewidth}{@{\extracolsep{\fill}}lrrr@{}}
\toprule
$k$ & \shortstack{Text\\OPD $-$ SFT} & \shortstack{VL\\OPD-RKL $-$ SFT} & \shortstack{VL\\FKL $-$ RKL} \\
\midrule
1 & $-10.00\;[-11.26, -8.76]$ & $+14.51\;[+13.27, +15.76]$ & $-0.26\;[-0.90, +0.39]$ \\
4 & $-8.40\;[-10.12, -6.66]$ & $+14.23\;[+12.74, +15.78]$ & $+0.20\;[-0.63, +1.06]$ \\
8 & $-5.23\;[-7.06, -3.36]$ & $+11.79\;[+10.28, +13.38]$ & $+0.34\;[-0.47, +1.22]$ \\
16 & $-2.31\;[-4.26, -0.33]$ & $+8.26\;[+6.75, +9.85]$ & $+0.48\;[-0.31, +1.34]$ \\
32 & $+0.09\;[-2.02, +2.23]$ & $+4.81\;[+3.31, +6.37]$ & $+0.37\;[-0.49, +1.30]$ \\
64 & $+2.25\;[-0.20, +4.69]$ & $+2.44\;[+0.88, +4.10]$ & $+0.00\;[-1.07, +1.17]$ \\
\bottomrule
\end{tabular*}
\end{table}

\subsection{Teacher Distributional Alignment}
\label{app:local-definitions}

The teacher probe compares student and teacher predictions at matching prefixes from each policy's own responses. At every retained position, it records the top 20 tokens and their log probabilities for both models. The resulting summaries reflect predictions at the contexts visited by that policy.

Let $U$ be the union of the two recorded token sets. We exponentiate the stored log probabilities, fill absent entries with zero, and normalize each vector separately over $U$ to obtain $\tilde p$ and $\tilde q$. With $m=(\tilde p+\tilde q)/2$, the recorded JS divergence is
\begin{equation}
\label{eq:js20}
\mathrm{JS}_{20}(\tilde p,\tilde q)
=\tfrac12\sum_{a\in U}\tilde p(a)\log\frac{\tilde p(a)}{m(a)}
+\tfrac12\sum_{a\in U}\tilde q(a)\log\frac{\tilde q(a)}{m(a)}.
\end{equation}

The implementation uses natural logarithms and adds $10^{-12}$ inside logarithms for numerical stability. This diagnostic compares truncated and renormalized predictions. OPD training itself uses full teacher distributions.

Top-1 agreement records whether the two models prefer the same token. Jaccard overlap divides the size of the intersection of their recorded candidate sets by the size of their union. Shared teacher mass sums the stored teacher probabilities on the intersection without renormalization. Each metric averages over positions available for both models, either overall or within the lexical categories defined below.

\begin{table}[H]
\centering
\small
\caption{Teacher alignment at prefixes generated by each policy. JS is defined in Eq.~\ref{eq:js20}; top-1 agreement and shared teacher mass are percentages. Position counts reflect response lengths and probe availability.}
\label{tab:app-local-compatibility}
\setlength{\tabcolsep}{3pt}
\renewcommand{\arraystretch}{1.08}
\begin{tabular*}{\linewidth}{@{\extracolsep{\fill}}lrrrrr@{}}
\toprule
Policy & Positions & JS $\downarrow$ & \shortstack{Top-1\\agreement} & Jaccard & \shortstack{Shared\\teacher mass} \\
\midrule
Text Base & 204,692 & 0.01701 & 93.58 & 0.5995 & 99.39 \\
Text SFT & 244,877 & 0.01720 & 94.25 & 0.6095 & 99.44 \\
Text OPD & 213,186 & 0.01468 & 93.82 & 0.6106 & 99.38 \\
\midrule
VL Base & 188,963 & 0.07734 & 81.89 & 0.5436 & 95.19 \\
VL SFT & 214,621 & 0.07208 & 83.88 & 0.5305 & 95.69 \\
VL OPD-RKL & 216,042 & 0.06431 & 83.67 & 0.5691 & 95.70 \\
VL OPD-FKL & 225,009 & 0.06234 & 83.49 & 0.5579 & 95.98 \\
\bottomrule
\end{tabular*}
\end{table}

The OPD checkpoints have lower JS than Base and SFT in both settings. SFT nevertheless has slightly higher top-1 agreement than OPD. Agreement on the preferred token and closeness of the surrounding distribution therefore capture different aspects of teacher transfer, motivating the distinction made in Sec.~\ref{sec:teacher-alignment}.

\subsection{Lexical Categories and Token Uncertainty}
\label{app:role-definitions}

Token entropy is aggregated with a fixed lexical classifier. The classifier strips surrounding whitespace from each decoded token string and assigns the first matching category in the order format, compute, strategy, and other. Empty strings and matches to the structural or chat-marker expression below are format tokens. Compute tokens consist entirely of digits or the characters \texttt{. + - * / = \string^ \% , ; :}. Strategy matching is case insensitive, uses word boundaries, and applies the following lexicon:
\begin{quote}
\small
let, assume, consider, suppose, if, then, else, therefore, hence, thus, since, because, so, but, however, alternatively, instead, notice, observe, note that, we can, we have, we get, we know, divide, multiply, substitute, factor, expand, simplify, approach, strategy, method, case, step.
\end{quote}
The category \texttt{strategy} identifies matches to this lexicon. Matching operates on individual decoded token strings, including for multiword entries, rather than reconstructed phrases. We use the term lexicon-conditioned entropy for this category throughout the analysis. The structural expression is given in Python raw-string notation:
\par\noindent\begin{minipage}{\linewidth}
\footnotesize
\begin{verbatim}
r"\\\\begin|\\\\end|\\\\boxed|\\\\frac|\\\\sqrt|\\\\left|\\\\right|"
r"\\\\text|\\\\mathbf|\\\\mathrm|\\\\cdot|\\\\times|\\\\div|"
r"\\\\leq|\\\\geq|\\\\neq|\\\\approx|\\\\equiv|\\\\infty|"
r"\\\\sum|\\\\prod|\\\\int|\\\\lim|\\\\log|\\\\ln|"
r"\\\\sin|\\\\cos|\\\\tan|"
r"\\\\\[|\\\\\]|\\\\\(|\\\\\)|\\\\{|\\\\}|\\\\&|\\\\\\\\|"
r"<think>|</think>|<\|im_start\|>|<\|im_end\|>"
\end{verbatim}
\end{minipage}\par

Entropy is computed from the truncated, renormalized candidate distributions in the separate one-response generation run. We average these estimates over matching token positions, with optional grouping by the correctness of the containing response. This weighting allows longer responses to contribute more positions. The categories provide a reproducible lexical summary of uncertainty, rather than labels for complete reasoning strategies.

\begin{table}[H]
\centering
\small
\caption{Token entropy at positions matching the reasoning lexicon. Values are in nats and average over matching token positions in the separate one-response run. Correctness refers to the containing response.}
\label{tab:app-breadth}
\setlength{\tabcolsep}{4pt}
\renewcommand{\arraystretch}{1.08}
\begin{tabular*}{\linewidth}{@{\extracolsep{\fill}}lrrr@{}}
\toprule
Policy & All responses & Correct responses & Incorrect responses \\
\midrule
Text Base & 0.971 & 0.998 & 0.966 \\
Text SFT & 0.242 & 0.820 & 0.202 \\
Text OPD & 1.048 & 1.170 & 1.031 \\
\midrule
VL Base & 0.622 & 0.762 & 0.573 \\
VL SFT & 0.214 & 0.287 & 0.196 \\
VL OPD-RKL & 0.629 & 0.675 & 0.605 \\
VL OPD-FKL & 0.952 & 0.950 & 0.954 \\
\bottomrule
\end{tabular*}
\end{table}

Standard RKL OPD retains entropy close to Base and above SFT in both settings. The same ordering relative to SFT appears within correct and incorrect responses. Together with the teacher-alignment measurements, this supports examining how teacher transfer changes the distribution over continuations beyond initial answer coverage.

\clearpage
% Trajectory source and objective are evaluated by performance after subsequent RL.
\section{Trajectory Sources and Objectives for RL Preparation}
\label{app:ablations}

The comparison in Sec.~\ref{sec:interventions} varies the trajectory source and KL objective while retaining the Qwen3-VL student and teacher, full teacher distributions, and the twenty-benchmark suite. This appendix reports the four-way comparison, diagnostics before RL, and complete OPD benchmark scores. Appendix~\ref{app:experimental-details} summarizes the VL OPD RKL/FKL configuration; Appendix~\ref{app:passk-definition} defines the reported Pass@$k$ intervals.

\subsection{Performance under Both Trajectory Sources}
\label{app:source-objective}

OPD applies teacher supervision at prefixes generated by the current student. TFD applies the same RKL or FKL loss at prefixes from teacher-generated responses. Both use full distributions, distinguishing the TFD variants from SFT with hard token targets. Table~\ref{tab:app-source-objective} supplies the scores plotted in Fig.~\ref{fig:intro}, using the same downstream RL configuration within the comparison.

\begin{table}[H]
\centering
\small
\caption{Performance before and after RL under both trajectory sources and KL objectives. Values average twenty benchmarks. Stage changes are computed from unrounded averages.}
\label{tab:app-source-objective}
\setlength{\tabcolsep}{4pt}
\begin{tabular*}{\linewidth}{@{\extracolsep{\fill}}llrrr@{}}
\toprule
Distillation prefixes & Objective & Before RL & After RL & Change \\
\midrule
Student (OPD) & RKL & 60.66 & 63.02 & $+2.36$ \\
Student (OPD) & FKL & 60.17 & 64.87 & $+4.70$ \\
Teacher (TFD) & RKL & 57.09 & 61.90 & $+4.82$ \\
Teacher (TFD) & FKL & 56.12 & 59.20 & $+3.08$ \\
\bottomrule
\end{tabular*}
\end{table}

Student rollouts yield higher scores than teacher rollouts under both objectives before and after RL. Within OPD, RKL leads at the distilled checkpoint, but FKL reaches the higher score after RL. These comparisons highlight the importance of both objective choice and the states receiving teacher supervision for subsequent learning.

\subsection{Coverage and Distributional Behavior before RL}
\label{app:token-support}

The OPD diagnostics connect the objective comparison to the analysis in Sec.~\ref{sec:support}. RKL has a Pass@1 of 48.06\%, slightly above FKL's 47.81\%, while both have a Pass@64 of 92.68\%. The paired intervals in Table~\ref{tab:passk_bootstrap} include zero at every reported $k$. These similar coverage measurements precede a final advantage of 1.85 points for FKL.

FKL has higher entropy at positions matching the reasoning lexicon, reaching 0.952 compared with 0.629 for RKL. It also has lower JS, at 0.06234 compared with 0.06431, while agreement on the teacher's preferred token is slightly lower. Tables~\ref{tab:app-breadth} and~\ref{tab:app-local-compatibility} report these quantities and the correctness splits. Shared teacher mass increases from 95.70\% to 95.98\%, while Jaccard overlap decreases from 0.5691 to 0.5579. The overlap of candidate sets and the probability assigned to those candidates therefore need not change together.

These observations are consistent with the hypothesis that closer teacher alignment and retained alternatives can help prepare a policy for RL. The trajectory-source comparison further shows that student rollouts yield higher final performance than teacher rollouts under both objectives, despite both sources receiving full teacher distributions. The evidence supports following each distilled checkpoint through RL when choosing the objective.

\clearpage
\subsection{Complete Benchmark Scores for the OPD Objectives}

The twenty-benchmark average for OPD-RKL increases from 60.6605 to 63.0215, a gain of 2.3610 points. OPD-FKL increases from 60.1670 to 64.8670, a gain of 4.7000 points. FKL begins 0.4935 points below RKL and ends 1.8455 points above it. Table~\ref{tab:app-ablation-benchmarks} reports all twenty task scores and the differences used in Fig.~\ref{fig:kl-sequence}.

\begin{table}[H]
\centering
\small
\caption{Complete scores after RL for OPD-RKL and OPD-FKL. Scores are percentages, and differences are percentage points. Averages are shown to four decimal places.}
\label{tab:app-ablation-benchmarks}
\setlength{\tabcolsep}{3pt}
\renewcommand{\arraystretch}{1.08}
\begin{tabular*}{\linewidth}{@{\extracolsep{\fill}}lrrr@{}}
\toprule
Benchmark & OPD-RKL & OPD-FKL & FKL $-$ RKL \\
\midrule
MMMU & 67.67 & 70.22 & $+2.55$ \\
MMMU-Pro & 54.71 & 55.38 & $+0.67$ \\
WeMath & 62.10 & 64.67 & $+2.57$ \\
Realworld-QA & 73.46 & 71.76 & $-1.70$ \\
MM-Star & 72.13 & 72.53 & $+0.40$ \\
SimpleVQA & 55.21 & 56.40 & $+1.19$ \\
MMIfeval & 63.55 & 63.16 & $-0.39$ \\
MMLongBench-Doc & 49.77 & 49.68 & $-0.09$ \\
AI2D & 85.62 & 84.94 & $-0.68$ \\
CharXiv-DQ & 89.05 & 88.60 & $-0.45$ \\
CharXiv-RQ & 49.20 & 52.20 & $+3.00$ \\
MMLU-Pro & 68.41 & 69.16 & $+0.75$ \\
SuperGPQA & 52.98 & 56.22 & $+3.24$ \\
AIME25 & 49.90 & 52.71 & $+2.81$ \\
AIME24 & 64.44 & 60.42 & $-4.02$ \\
AIME26 & 58.13 & 59.79 & $+1.66$ \\
AMC23 & 87.27 & 91.88 & $+4.61$ \\
AMC24 & 62.29 & 70.14 & $+7.85$ \\
HMMT & 31.04 & 35.52 & $+4.48$ \\
OlympiadBench & 63.50 & 71.96 & $+8.46$ \\
\midrule
Average & 63.0215 & 64.8670 & $+1.8455$ \\
\bottomrule
\end{tabular*}
\end{table}

FKL finishes higher on 14 of twenty tasks. The largest gains include 8.46 points on OlympiadBench and 7.85 points on AMC24, while AIME24 is 4.02 points lower. The final average therefore summarizes a broad but nonuniform benefit across tasks.

\clearpage
% Diagnostic scoring is documented separately from the metric definitions.
\section{Diagnostic Scoring}
\label{app:response-processing}

This appendix specifies diagnostic scoring. The coverage, teacher-alignment, and token-uncertainty measurements are defined in Appendix~\ref{app:metrics}.

Each policy has 64 responses for each of 1,024 diagnostic prompts, giving 65,536 responses. Responses are generated in non-thinking mode. The answer extractor first attempts the last boxed expression, then supported answer labels, natural-language answer statements, and multiple-choice letters. Correctness is checked against the reference through normalized exact matching and symbolic equivalence, with an LLM judge used for cases requiring further adjudication.

\clearpage
\endgroup

\end{document}